\documentclass[letterpaper]{article} 
\usepackage{geomoe}
\nocopyright
\usepackage{times}  
\usepackage{helvet}  
\usepackage{courier}  
\usepackage[hyphens]{url}  
\usepackage{graphicx} 
\usepackage{natbib}  
\usepackage{caption} 
\usepackage{algorithm}
\usepackage{algorithmic}
\usepackage{amsmath}
\usepackage{amsfonts}
\usepackage{multirow}
\usepackage{booktabs}
\usepackage{arydshln}
\usepackage{mathrsfs}
\usepackage{newfloat}

\usepackage{listings}
\usepackage[hyphens]{url}
\DeclareCaptionStyle{ruled}{labelfont=normalfont,labelsep=colon,strut=off}
\floatstyle{ruled}
\newfloat{listing}{tb}{lst}{}
\floatname{listing}{Listing}

\title{GeoMoE: Divide-and-Conquer Motion Field Modeling with Mixture-of-Experts for Two-View Geometry}
\author{
	Jiajun Le, 
	Jiayi Ma\thanks{Corresponding authors.}
}
\affiliations{
	Wuhan University, China \\
	jiajunle01@gmail.com, jyma2010@gmail.com
}

\begin{document}
	\maketitle
	
	\begin{abstract}
		Recent progress in two-view geometry increasingly emphasizes enforcing smoothness and global consistency priors when estimating motion fields between pairs of images. However, in complex real-world scenes, characterized by extreme viewpoint and scale changes as well as pronounced depth discontinuities, the motion field often exhibits diverse and heterogeneous motion patterns. Most existing methods lack targeted modeling strategies and fail to explicitly account for this variability, resulting in estimated motion fields that diverge from their true underlying structure and distribution. We observe that Mixture-of-Experts (MoE) can assign dedicated experts to motion sub-fields, enabling a divide-and-conquer strategy for heterogeneous motion patterns. Building on this insight, we re-architect motion field modeling in two-view geometry with GeoMoE, a streamlined framework. Specifically, we first devise a Probabilistic Prior-Guided Decomposition strategy that exploits inlier probability signals to perform a structure-aware decomposition of the motion field into heterogeneous sub-fields, sharply curbing outlier-induced bias. Next, we introduce an MoE-Enhanced Bi-Path Rectifier that enhances each sub-field along spatial-context and channel-semantic paths and routes it to a customized expert for targeted modeling, thereby decoupling heterogeneous motion regimes, suppressing cross-sub-field interference and representational entanglement, and yielding fine-grained motion-field rectification. With this minimalist design, GeoMoE outperforms prior state-of-the-art methods in relative pose and homography estimation and shows strong generalization.
		The source code and pre-trained models are available at \url{https://github.com/JiajunLe/GeoMoE}.
	\end{abstract}

	\section{Introduction}
	
	Establishing precise and reliable correspondences between two images remains a critical and enduring challenge in computer vision, underpinning a variety of advanced vision tasks such as image stitching~\cite{Brown2007Automatic}, Simultaneous Localization and Mapping~\cite{MurArtal2015ORBSLAM}, and Structure from Motion~\cite{schonberger2016structure}. Typically, these correspondences are inferred based on the similarity between local descriptors (e.g., SIFT~\cite{Lowe2004SIFT}, SuperPoint~\cite{DeTone2018SuperPoint}) extracted by detectors. Nevertheless, in complex, real-world scenarios characterized by significant viewpoint changes and illumination variations, the discriminative capability of these descriptors is often severely compromised, resulting in numerous incorrect matches (i.e., outliers). Consequently, such unreliable matches undermine the accuracy and robustness required for subsequent geometric estimation tasks. Therefore, effectively rejecting outliers while preserving accurate correspondences (i.e., inliers) has emerged as an essential research issue, which is the focus of this paper.
	
	\begin{figure}[t]
		\centering
		\includegraphics[width=1\linewidth]{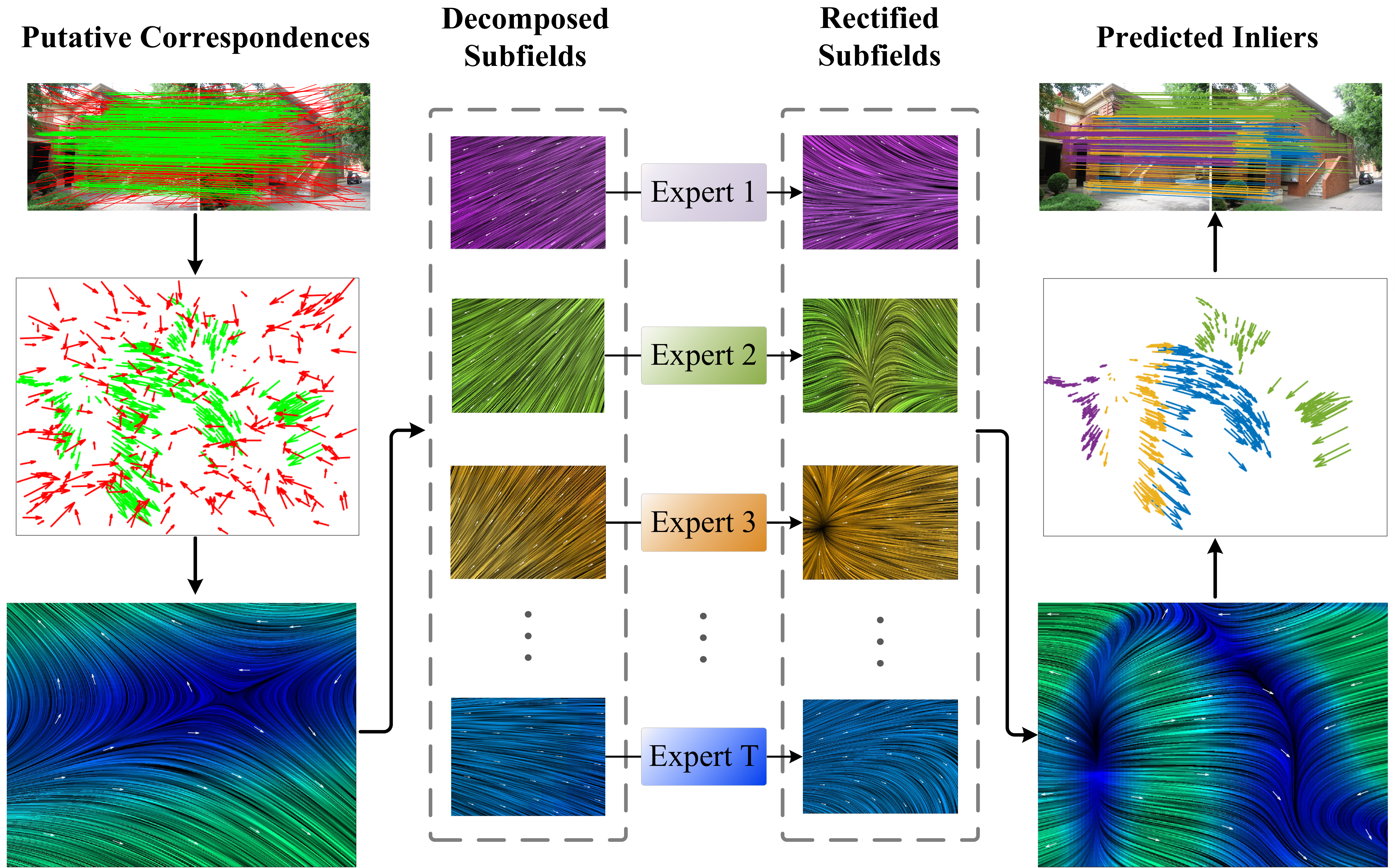}
		\caption{Illustration of MoE modeling heterogeneous motion patterns.
			Red lines denote outliers, while green lines represent inliers.
			Different colors in the filtered motion field indicate distinct motion sub-fields, each corresponding to a specific heterogeneous motion pattern.}
		\label{head}
	\end{figure}
	
	The rapid advances in deep learning over recent years have significantly accelerated research into two-view geometry problems. PointCN~\cite{yi2018learning} serves as a flagship method, employing an MLP backbone to recast outlier rejection as a binary classification task distinguishing outliers from inliers. Successor methods derived from PointCN, such as OANet~\cite{zhang2019learning}, NCMNet~\cite{liu2023progressive}, U-Match~\cite{li2023u}, and BCLNet~\cite{ Miao2024BCLNet}, have continued to demonstrate consistently improved performance. Another line of work takes a more natural and structured perspective by leveraging consistency and smoothness priors that arise from the uniform deformation between image pairs.
	Recent methods, including3 LMCNet~\cite{liu2021learnable}, ConvMatch++~\cite{Zhang2023ConvMatch}, DeMatch~\cite{zhang2024dematch}, and DeMo~\cite{Lu2025DeMo}, explicitly exploit these priors, validating their effectiveness in improving model performance. Nevertheless, these methods are still subject to notable limitations. Specifically, LMCNet enforces motion consistency via Laplacian regularization but incurs high computational cost due to expensive matrix decomposition. ConvMatch++ structures unordered motion fields into image-like forms for CNN-based context extraction, but faces local over-smoothing and lacks global perceptual awareness. DeMo models global consistency within an RKHS framework~\cite{aronszajn1950theory, baldassarre2012multi}, while effectively preserving motion discontinuities while mitigating the risk of over-smoothing.
	Nonetheless, in realistic imaging conditions, extreme visual scenarios, drastic scale variations, and significant depth disparities often cause motion fields to exhibit multiple heterogeneous motion patterns as shown in Fig.~\ref{head}. A straightforward strategy to address this is to decompose the motion field into sub-fields, as in DeMatch. However, DeMatch applies a uniform modeling strategy to all sub-fields, inevitably overlooking the intrinsic variability and distinct motion characteristics of each. Consequently, an unresolved key issue persists in the literature: \emph{given the pronounced differences in both feature characteristics and distributions across motion sub-fields, how can we develop tailored, adaptive modeling strategies to effectively address their inherent heterogeneity?}
	
	Rather than relying on complex and redundant handcrafted designs, we advocate Mixture-of-Experts (MoE)~\cite{Jacobs1991Adaptive} as a natural and efficient alternative. In this framework, a routing network dynamically analyzes the motion characteristics of each sub-field and assigns it to the most suitable expert for dedicated modeling. It allows experts to specialize in distinct motion patterns, preserving the independence of the modeling process and yielding more discriminative features for accurate sub-field rectification.
	
	Building upon this insight, we propose GeoMoE, a novel and streamlined network paradigm aimed at rebuilding motion field refinement for two-view geometry through the lens of Mixture-of-Experts.
	Specifically, we heuristically design a Probabilistic Prior-Guided Decomposition strategy that leverages inlier probabilities from the preceding layer as priors to guide a soft, structure-aware clustering of the motion field. This process decomposes the motion field into multiple sub-fields with strong scene-level consistency by concentrating on regions exhibiting similar motion patterns, while concurrently mitigating bias caused by uneven outliers.
	Upon completing the decomposition, we propose a novel MoE-Enhanced Bi-Path Rectifier, which first comprises two complementary paths: a spatial path that captures relative motion cues within the spatial context of each sub-field, and a channel path that enhances feature representations along the semantic dimension. A routing network then assigns each sub-field to the suitable expert based on its feature characteristics and distributions, enabling each expert to specialize in modeling a specific motion pattern. This targeted expert assignment alleviates cross-sub-field information interference and representational entanglement, thereby yielding more precise tailored sub-field rectification. Through synergistic integration, the rectified sub-fields collectively reconstruct a global motion field with enriched detail, enhanced structural clarity, and closer alignment to real-world distributions.
	
	Taken as a whole, as illustrated in Fig.~\ref{head}, our method forms a unified and streamlined pipeline from decomposition to expert-guided modeling, fully leveraging MoE to address heterogeneous motion fields.  Our main contributions are summarized as follows:
	\begin{itemize}
		\item We propose GeoMoE, a groundbreaking two-view geometry network that introduces MoE to enable fine-grained modeling and rectification of motion sub-fields.
		\item We develop a probabilistic prior-guided decomposition strategy that utilizes inlier probability as a prior to perform structure-aware motion field decomposition, while reducing bias induced by uneven outliers.
		\item We architect an innovative MoE-enhanced bi-path rectifier that reduces mutual interference and entanglement among sub-fields with diverse motion patterns, facilitating expert-driven, fine-grained modeling and rectification across sub-fields.
		\item Extensive experiments across multiple public datasets show that GeoMoE surpasses state-of-the-art methods, validating its effectiveness and strong generalizability.
	\end{itemize}

	\begin{figure*}[t]
		\centering
		\includegraphics[width=1\linewidth]{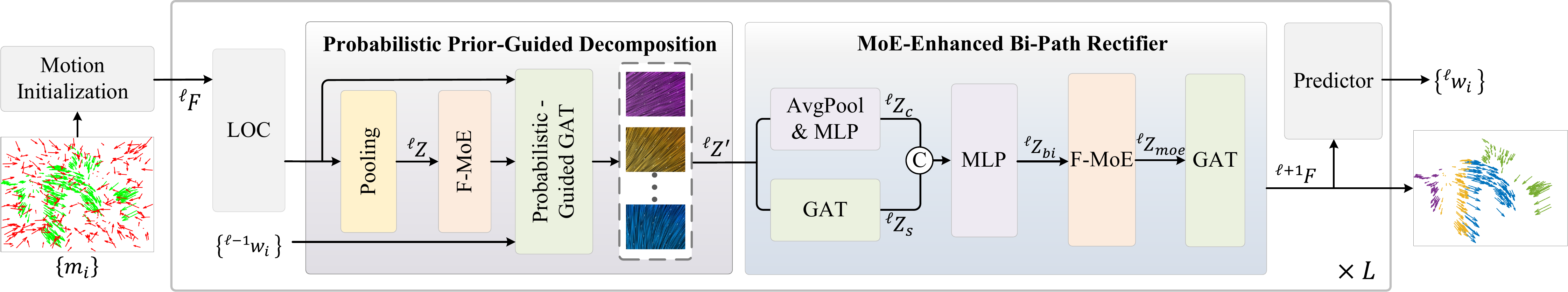}
		\caption{Overall architecture of GeoMoE.
			Following initialization and the LOC module, high-dimensional motion features $\sideset{^{\ell}}{}{\mathop{\boldsymbol{F}}}$ are decomposed into multiple motion sub-fields $\sideset{^\ell}{}{\mathop{\boldsymbol{Z}^{\prime}}}$ via the probabilistic prior-guided decomposition strategy. In the MoE-enhanced bi-path rectifier, each sub-field undergoes bi-path feature enhancement and subsequently is modeled by customized experts within F-MoE. The complete motion field $\sideset{^{\ell+1}}{}{\mathop{\boldsymbol{F}}}$ is then reconstructed by a GAT module. Finally, the inlier probabilities $\sideset{^{\ell}}{}{\mathop{\boldsymbol{w}}}$ predicted at each layer are utilized in loss computation.}
		\label{Architecture}
	\end{figure*}
	
	\section{Related Work}
	\subsection{Learning-Based Outlier Rejection Methods}
	PointCN~\cite{yi2018learning} pioneered the application of deep learning to the outlier rejection task by reframing it as an inlier/outlier classification problem modeled with MLPs. A series of subsequent works have focused on enhancing feature representations through increasingly sophisticated architectural designs. For example, OANet~\cite{zhang2019learning} proposes differentiable pooling to capture local geometry, MS$^2$DG-Net~\cite{Dai2022MS2DGNet} introduces a graph-based representation in feature space to model inter-correspondence relationships, and U-Match~\cite{li2023u} designs a U-shaped architecture to enable hierarchical and multi-scale local context learning. Furthermore, CLNet~\cite{zhao2021progressive} establishes a local-to-global consistency learning paradigm, progressively preserving reliable matches. This progressive filtering paradigm has been adopted by NCMNet~\cite{liu2023progressive}, BCLNet~\cite{Miao2024BCLNet}, and VSFormer~\cite{Liao2024VSFormer}.
	
	From another perspective, ConvMatch++~\cite{Zhang2023ConvMatch} reinterprets two-view geometry as a motion field estimation problem, structuring unordered correspondences into fields and applying CNNs for local consistency. DeMatch~\cite{zhang2024dematch} alleviates computational burden by decomposing the global motion field into multiple implicit sub-fields, while DeMo~\cite{Lu2025DeMo} introduces learnable deep kernels in RKHS to capture global motion coherence. Nevertheless, these methods fail to account for multiple heterogeneous motion patterns induced by complex scene variations, which pose a significant challenge to precise motion field modeling.
	
	\subsection{Mixture-of-Experts (MoE)}
	The Mixture-of-Experts (MoE)~\cite{Jacobs1991Adaptive} framework leverages a routing network to dynamically activate specialized expert networks, enabling task-adaptive representation learning. It has shown strong effectiveness in both vision~\cite{Cai2018ContrastEnhancer} and language models~\cite{Lepikhin2020GShard}, particularly when integrated into Transformer~\cite{Fedus2022Switch} to boost capacity and efficiency. Recent work extends MoE to various tasks: it reconfigures U-Net skip connections for semantic decoding in medical segmentation~\cite{Luo2025RethinkingUNet}; assigns correspondences to shared experts in point cloud registration~\cite{Huang2025PSReg} to enhance feature discriminability in overlapping regions; and improves robustness under cross-domain variations in correspondence pruning~\cite{Xia2025CorrMoE}.
	In contrast, we introduce MoE into two-view geometry to address heterogeneous motion patterns arising from scene diversity and complexity at the underlying correspondence level.
	
	\section{Method}
	\subsection{Problem Formulation}
	Following prior work~\cite{liu2021learnable}, given a pair of images $(\mathbf{I}, \mathbf{I}^{\prime})$, we use an off-the-shelf descriptor extraction method~\cite{Lowe2004SIFT} to detect keypoints and compute descriptors. Based on descriptor similarity, a putative correspondence set
	$\boldsymbol{C}=\{\boldsymbol{c}_i|\boldsymbol{c}_i=(\boldsymbol{x}_i,\boldsymbol{x}_i^{\prime})\} \in \mathbb{R}^{N\times 4}$ is constructed using a nearest neighbor (NN) matcher, where $\boldsymbol{x}_i$ and $\boldsymbol{x}_i^{\prime}$ denote the keypoint coordinates normalized by the respective camera intrinsics. Each correspondence $\boldsymbol{c}_i$ defines a motion vector $\boldsymbol{m}_i=(\boldsymbol{x}_i,\boldsymbol{x}_i^{\prime}-\boldsymbol{x}_i)$, which is then projected into a high-dimensional feature $\boldsymbol{f}_i$ to form the dense motion field $\boldsymbol{F}=\{\boldsymbol{f}_i\} \in \mathbb{R}^{N\times D}$. After the locality orthogonal context module, we apply a probabilistic prior-guided decomposition strategy to decompose the global motion field into multiple structure-aware sub-fields. Each sub-field, enhanced along spatial-context and channel-semantic paths, is subsequently routed by the MoE-enhanced bi-path rectifier to a customized expert for dedicated modeling and refinement. The rectified sub-fields are then integrated to reconstruct a high-fidelity motion field with fine-grained structure.
	A lightweight MLP then predicts an inlier/outlier probability $w_i$ for each correspondence.
	This cycle of local context aggregation, decomposition, refinement, and reconstruction is repeated across $L$ sequential layers, progressively enhancing the motion field estimation. Finally, a weighted eight-point algorithm, guided by the predicted probabilities $\boldsymbol{w}=\{w_i\} \in \mathbb{R}^{N\times 1}$, is used to estimate the essential matrix $\hat{\boldsymbol{E}}$.
	The overall process is formalized as:
	\begin{equation}
		\begin{aligned}
			\boldsymbol{w} &= \boldsymbol{g}_{\phi}(\boldsymbol{C}),  \\
			\hat{\boldsymbol{E}} &= \boldsymbol{h}(\boldsymbol{w}, \boldsymbol{C}),
			\label{Formulation}
		\end{aligned}
	\end{equation}
	where $\boldsymbol{g}_{\phi}$ denotes GeoMoE parameterized by $\phi$, and $\boldsymbol{h}$ represents the robust weighted eight-point algorithm. As shown in Fig.~\ref{Architecture}, the overall architecture of GeoMoE is depicted, and we elaborate on its design in the following section.
	
\subsection{Motion Initialization}
	Owing to their low dimensionality, motion vectors ${\boldsymbol{m}_i}$ lack the expressive power needed to capture deep features. To overcome this, inspired by prior work~\cite{Lu2025DeMo}, we embed $\boldsymbol{m}_i$ into a high-dimensional space to obtain a richer representation $\boldsymbol{f}_i \in \mathbb{R}^{D}$ that serves as input to the first layer:
	\begin{equation}
		\sideset{^{0}}{_i}{\mathop{\boldsymbol{f}}}=\mathcal{E}{(\boldsymbol{m}_i)},\quad i=1,\dots,N.
		\label{MotionEmbedding}
	\end{equation}
	Here, $\mathcal{E}(\cdot)$ denotes a transformation that elevates motion vectors from low- to high-dimensional space, implemented via a convolutional layer followed by ContextNorm~\cite{yi2018learning}. This embedding improves feature expressivity and lays a robust foundation for downstream processing.

	\subsection{Mixture-of-Experts for Motion Field (F-MoE)}
\begin{figure}[t]
		\centering
		\includegraphics[width=1\linewidth]{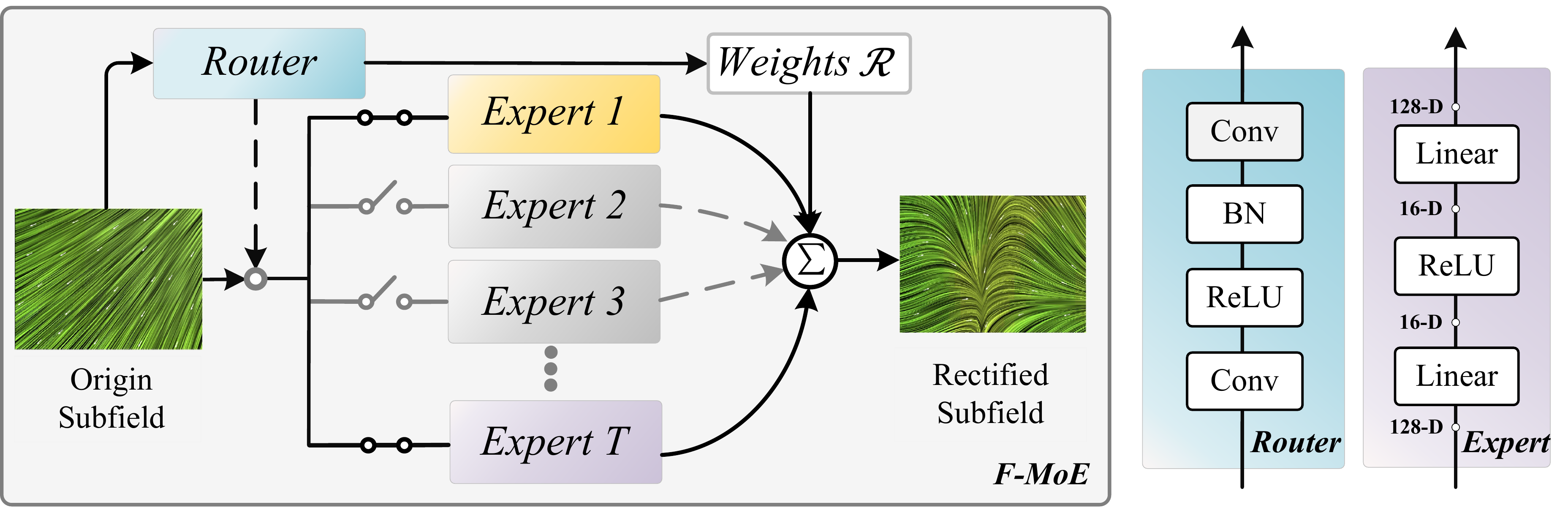}
		\caption{Workflow diagram of F-MoE.}
		\label{moe}
	\end{figure}

	The Mixture-of-Experts for sub-field modeling comprises a routing network and a set of feed-forward expert networks, as shown in Fig.~\ref{moe}. Given sub-field inputs $\boldsymbol{f}_i$, the routing network dynamically activates the most relevant experts for specialized modeling. Unlike conventional fully-connected routing, we employ a sparse MLP to better reflect the sparsity and region-specific nature of sub-field rectification. The routing network is defined as:
	\begin{equation}
		\mathcal{R}=\text{top-k}(\text{softmax}(\text{MLP}(\boldsymbol{f}_i))),
		\label{Routing}
	\end{equation}
	where the \text{MLP} shares parameters across all sub-fields, and the \text{top-k} operator selects the $k$ most responsive experts based on activation scores. Each expert is a lightweight feed-forward network, tailored for dimensionality reduction and targeted motion field refinement. The final output is a weighted summation of the activated experts’ outputs:
	\begin{equation}
		\boldsymbol{f}_i^{\prime}=\sum_{j\in \mathbb{N}_k} \mathcal{R}_j \cdot E_j(\boldsymbol{f}_i),
		\label{MoE}
	\end{equation}
	where $\mathbb{N}_k$ denote the indices of the \text{top-k} experts, $E_j$ denotes the $j$-th expert, and $\mathcal{R}_j$ is its routing weight.

	\subsection{Probabilistic Prior-Guided Decomposition}
	The motion field derived from putative correspondences often manifests a spectrum of heterogeneous dynamic patterns due to the complexity of real-world scenes, such as viewpoint changes, scale variations, and pronounced depth discontinuities. Instead of modeling the field holistically, prior work~\cite{zhang2024dematch} shows that grouping motion vectors with coherent geometric behavior yields a more efficient alternative.
	However, prior approaches that rely on learnable tokens to implicitly model motion patterns tend to neglect explicit spatial priors, i.e., the natural inductive bias that spatially adjacent motions are likely to share consistent transformations, resulting in physically ambiguous and semantically uninterpretable representations.
	
	To address these limitations, we propose a Probabilistic Prior-Guided Decomposition (PPGD) strategy rooted in the structural regularities of the scene. Specifically, we first apply a pooling operation~\cite{zhang2019learning} to coarsely partition the motion field $\sideset{^\ell}{}{\mathop{\boldsymbol{F}}}=\{\sideset{^\ell}{}{\mathop{\boldsymbol{\boldsymbol{f}}_{i}}}\}_{i=1}^{N}$ into $M$ preliminary sub-fields $\sideset{^\ell}{}{\mathop{\boldsymbol{Z}}} \in \mathbb{R}^{M\times D}$:
	\begin{equation}
		\begin{aligned}
			\sideset{^\ell}{}{\mathop{\boldsymbol{Z}}}&=\boldsymbol{S}\sideset{^\ell}{}{\mathop{\boldsymbol{F}}},\\
			\boldsymbol{S}&=\text{softmax}(\text{MLP}(\sideset{^\ell}{}{\mathop{\boldsymbol{F}}})).
		\end{aligned}
		\label{Decomposition}
	\end{equation}
	This operation is inherently structure-aware, leveraging localized contextual cues to inform sub-field assignment, as demonstrated in prior work~\cite{zhang2019learning}. To enhance the expressiveness and attenuate distortions caused by outliers, we incorporate an F-MoE module $\mathcal{M}(\cdot)$ that performs specialized refinement, guided by a probabilistic masking strategy using the previous-layer confidence scores $\sideset{^{\ell-1}}{}{\mathop{\boldsymbol{w}}}=\{\sideset{^{\ell-1}}{}{\mathop{w}_i}\} \in \mathbb{R}^{N}$:
	\begin{equation}
		\sideset{^\ell}{}{\mathop{\boldsymbol{Z}^{\prime}}}=\mathcal{G}(\mathcal{M}(\sideset{^\ell}{}{\mathop{\boldsymbol{Z}}}), \sideset{^{\ell-1}}{}{\mathop{\boldsymbol{W}}}\odot\sideset{^\ell}{}{\mathop{\boldsymbol{F}}}),\\
		\label{ProGuide}
	\end{equation}
	where $\sideset{^{\ell-1}}{}{\mathop{\boldsymbol{W}}}\in \mathbb{R}^{N \times D}$ is a broadcasted version of $\sideset{^{\ell-1}}{}{\mathop{\boldsymbol{w}}}$, $\odot$ denotes the Hadamard product, and $\mathcal{G}(\cdot)$ refers to the Graph Attention Network~\cite{Velickovic2017GAT}. For the initial layer, $\sideset{^0}{}{\mathop{\boldsymbol{w}}}$ is initialized to an all-ones vector, implying no prior suppression.
	This prior-informed decomposition strategy is capable of yielding a motion field that is both physically plausible and structurally coherent.

	\begin{table*}[t]
		\centering
		\resizebox{0.9\linewidth}{!}{
			\setlength{\tabcolsep}{5mm}{
				\begin{tabular}{lccccccc}
					\toprule
					\multirow{2}{*}{Method} & \multicolumn{3}{c}{YFCC100M (\%$\uparrow$)}
					&& \multicolumn{3}{c}{SUN3D (\%$\uparrow$)}\\
					\cmidrule{2-4} \cmidrule{6-8}
					& $@5^\circ$     & $@10^\circ$     & $@20^\circ$
					&& $@5^\circ$     & $@10^\circ$     & $@20^\circ$ \\
					\midrule
					RANSAC     & 3.47       &9.10      & 18.60
					&& 1.04        & 3.43      & 8.75     \\
					MAGSAC++     &  1.51       & 4.67      &11.95
					&&  0.88		  & 2.99      & 8.42	   \\
					NG-RANSAC     &  16.47       & 28.38      &42.03
					&&  4.22		  & 10.87      & 21.39	   \\
					\hdashline
					GMS     & - / 13.29       & - / 24.38      & - / 37.83
					&& - / 4.12        & - / 10.53      & - / 20.82     \\
					VFC     & - / 17.43       & - / 29.98      & - / 43.00
					&& - / 5.26		  & - / 13.05      & - / 24.84	   \\
					\hdashline
					PointCN & 10.16 / 26.73 & 24.43 / 44.01 & 43.31 / 60.49
					&& 3.05 / 6.09   & 10.00 / 15.43 & 24.06 / 29.74    \\
					OANet & 15.92 / 27.26 & 35.93 / 45.93 & 57.11 / 63.17
					&& 5.93 / 6.78   & 16.91 / 17.10 & 34.32 / 32.41    \\
					NCMNet & 26.89 / 32.30 & 46.19 / 52.29 & 64.21 / 69.65
					&& 6.31 / 7.10   & 16.84 / 18.56 & 33.11 / \textbf{35.58} \\
					UMatch & 30.93 / 33.92 & 52.17 / 53.09 & 69.75 / 69.45
					&& 8.03 / 7.01   & 20.83 / 17.79 & 38.75 / 33.57     \\
					MS$^2$DG-Net & 20.61 / 31.55 & 42.90 / 50.94 & 64.26 / 68.34
					&& 5.89 / 6.95   & 16.85 / 17.66 & 34.25 / 33.38  \\
					DeMatch & 30.89 / 32.98 & 52.67 / 52.37 & 70.33 / 69.01
					&& \underline{9.31} / 7.44   & 23.10 / 18.66 & 41.55 / 34.78  \\		
					DeMo & \underline{32.57} / \textbf{34.63} & \underline{54.82} / \underline{54.42} & \underline{72.51} / \underline{71.15}
					&& \underline{9.31} / \underline{7.48}   & \underline{23.18} / \underline{18.85} & \underline{41.75} / \underline{35.31} \\	
					GeoMoE (Ours) & \textbf{34.21} / \underline{34.52} & \textbf{56.25 / 54.72} & \textbf{73.53 / 71.63}
					&& \textbf{9.98 / 7.54}   & \textbf{24.05 / 18.87} & \textbf{42.71} / 35.22 \\	
					\bottomrule
		\end{tabular}}}
		\vspace{-0.05in}
		\caption{Quantitative results of relative pose estimation. AUC is reported under $5^\circ$, $10^\circ$, and $20^\circ$ thresholds with the weighted eight-point algorithm / RANSAC. \textbf{Best} and \underline{second-best} results are shown in bold and underlined, respectively.}
		\label{table1:Relative Pose Estimation}
	\end{table*}

	\subsection{MoE-Enhanced Bi-Path Rectifier}
	Amid substantial heterogeneity in motion patterns across sub-fields,  prior approaches~\cite{zhang2024dematch} that adopt a shared architecture for all sub-fields inherently fail to capture the diverse statistical characteristics and spatial dynamics specific to each region. This uniform treatment overlooks crucial structural discrepancies, resulting in constrained representational capacity and limited modeling fidelity.
	
	To address this, we propose the MoE-enhanced Bi-Path Rectifier (MBPR), a specialized rectification module that routes each sub-field to a set of dedicated expert networks for adaptive refinement and motion field rectification, enabling more tailored and expressive modeling.
	Specifically, we begin by enriching sub-field representations through a bi-path encoding strategy:
	the spatial branch models relative motion via graph-based aggregation, while the channel branch extracts high-level semantics by applying global average pooling followed by channel-wise modulation. The two branches are defined as:
	\begin{equation}
		\begin{aligned}
			\sideset{^\ell}{}{\mathop{\boldsymbol{Z}_{s}}}&=\mathcal{G}(\sideset{^\ell}{}{\mathop{\boldsymbol{Z}^{\prime}}}, \sideset{^\ell}{}{\mathop{\boldsymbol{Z}^{\prime}}}),\\
			\sideset{^\ell}{}{\mathop{\boldsymbol{Z}_{c}}}&=\text{MLP}(\text{AvgPool}( \sideset{^\ell}{}{\mathop{\boldsymbol{Z}^{\prime}}}))\sideset{^\ell}{}{\mathop{\boldsymbol{Z}^{\prime}}}.\\
		\end{aligned}
		\label{bi_path}
	\end{equation}
	The outputs are concatenated and fused via a lightweight MLP to obtain the final bi-path enhanced representation:
	\begin{equation}
		\sideset{^\ell}{}{\mathop{\boldsymbol{Z}_{bi}}}=\text{MLP}( \sideset{^\ell}{}{\mathop{\boldsymbol{Z}_{c}}}||\sideset{^\ell}{}{\mathop{\boldsymbol{Z}_{s}}}).\\
		\label{cat}
	\end{equation}
	Next, each enhanced sub-field is dynamically assigned to a subset of expert networks via a sparsely-activated routing mechanism:
	\begin{equation}
		\sideset{^\ell}{}{\mathop{\boldsymbol{Z}_{moe}}}=\mathcal{M}( \sideset{^\ell}{}{\mathop{\boldsymbol{Z}_{bi}}}),\\
		\label{MoEre}
	\end{equation}
	where $\mathcal{M}(\cdot)$ denotes the F-MoE module that adaptively assigns sub-fields to specialized experts according to their individual characteristics.
	Finally, the rectified sub-fields are globally aggregated to reconstruct the updated motion field:
	\begin{equation}
		\sideset{^{\ell+1}}{}{\mathop{\boldsymbol{F}}}=\mathcal{G}( \sideset{^{\ell}}{}{\mathop{\boldsymbol{F}}},\sideset{^\ell}{}{\mathop{\boldsymbol{Z}_{moe}}}).\\
		\label{restruction}
	\end{equation}
	This staged rectification pipeline progressively refines the global motion representation with enhanced structural clarity, higher spatial fidelity, and stronger alignment to the true underlying motion distribution.

	\subsection{Local Orthogonal Context (LOC)}
	Before motion field decomposition and sub-field modeling, we incorporate orthogonal contextual cues within KNN-based local neighborhoods, following~\cite{Lu2025DeMo}. This preparatory step facilitates the detection of local motion discontinuities while suppressing excessive smoothness artifacts.
	Concretely, each motion vector $\boldsymbol{f}_i$ undergoes dimensionality reduction. A KNN-based neighborhood $\{\boldsymbol{f}_{i,j}\}_{j=0}^{K}$ is then constructed to capture its local structure, and pairwise differences $\boldsymbol{h}_{i,j}=\boldsymbol{f}_{i}-\boldsymbol{f}_{i,j}$ are computed.
	To jointly capture spatial correlations and feature-level semantics, we utilize a two-stage aggregation scheme composed of lightweight MLPs:
	\begin{equation}
		\begin{aligned}
			\boldsymbol{H}_{f}&=\text{MLP}(\boldsymbol{H})+\boldsymbol{H},\\
			\boldsymbol{F}_{local}&=\text{MLP}(\text{MLP}(\boldsymbol{H}_{f}^{T})+\boldsymbol{H}_{f}^{T}).\\
		\end{aligned}
		\label{local}
	\end{equation}
	The final output $\boldsymbol{F}_{local}$ is projected into a higher-dimensional space, equipping it with enriched local representations for downstream motion modeling.

	\subsection{Loss Functions}
	During supervised training, we adopt the widely used binary cross-entropy loss for inlier/outlier classification, together with a regression loss for essential matrix estimation. Importantly, to address the often-overlooked issue of expert imbalance in the Mixture-of-Experts (MoE) framework~\cite{Fedus2022Switch}, we design a load-balancing regularization term:
	\begin{equation}
		\mathcal{L}_{load}=\frac{1}{T}\sum_{t=1}^{T}(\frac{1}{N}\sum_{i=1}^{N} \sideset{^{\ell}}{}{\mathop{\boldsymbol{r}_{i,t}}})^{2},
		\label{loadloss}
	\end{equation}
	where $\sideset{^{\ell}}{}{\mathop{\boldsymbol{r}_{i,t}}}$ denotes the activation probability of the $t$-th
	expert for the $i$-th sub-field, and $T$ is the total number of experts. This regularization encourages balanced expert utilization, promoting more stable training and better generalization. The overall loss function is defined as:
	\begin{equation}
		\mathcal{L}=\sum_{\ell=0}^{L-1}\mathcal{L}_{cls}\left(\boldsymbol{w},\sideset{^{\ell}}{}{\mathop{\boldsymbol{w}}}\right)+\mu\mathcal{L}_{reg}\left(\mathbf{E},\sideset{^{\ell}}{}{\mathop{\hat{\mathbf{E}}}}\right)+\frac{\beta}{L}\mathcal{L}_{load},
		\label{loss}
	\end{equation}
	where $\beta$ and $\mu$ are weighting coefficients that balance the contributions of each term.
	
	\subsection{Implementation Details}
	In our implementation, GeoMoE is unrolled for $L=8$ layers, with motions represented in a $128$-dimensional space (i.e., $D=128$). For each image pair, up to $N=2000$ putative matches are obtained using SIFT followed by NN matcher.
	The motion field is decomposed into $48$ sub-fields. Each F-MoE comprises $4$ experts, and every sub-field is routed to $2$ of them for parallel processing.
	For loss balancing, the regularization weight $\mu$ is initially set to $0$ and increased to $0.5$ after $20k$ iterations. The weight $\beta$ for expert load balancing remains fixed at $0.01$.
	The training details are provided in the Supplementary Material (\textit{S.M.}).

	\section{Experiments}
	
	\subsection{Relative Pose Estimation}
	Many vision applications require estimating the relative camera pose, i.e., rotation and translation, between image pairs from the same scene, thereby imposing stringent demands on high-quality inliers.
	
\subsubsection{Comparisons with Outlier Rejection Methods} 
	\noindent{\textit{Datasets and Evaluation Protocols:}}
	Following the standard evaluation protocol~\cite{zhang2019learning}, we conduct relative pose estimation experiments on the outdoor dataset YFCC100M~\cite{thomee2016yfcc100m} and the indoor dataset SUN3D~\cite{xiao2013sun3d}. We report the Area Under the cumulative error Curve (AUC) for the maximum angular error in both rotation and translation under multiple thresholds.
	More detailed experimental settings are provided in the \textit{S.M}.

	\noindent{\textit{Baselines:}}
	We benchmark our method against a broad range of baselines, including both classical methods (GMS~\cite{bian2017gms}, and VFC~\cite{ma2014robust}) and deep learning-based approaches (PointCN~\cite{yi2018learning}, OANet~\cite{zhang2019learning}, NCMNet~\cite{liu2023progressive}, U-Match~\cite{li2023u}, MS$^2$DG-Net~\cite{Dai2022MS2DGNet}, DeMatch~\cite{zhang2024dematch}, and DeMo~\cite{Lu2025DeMo}).
	In addition, we evaluate against the RANSAC family of methods, namely RANSAC~\cite{fischler1981random}, MAGSAC++~\cite{Barath2020MAGSAC}, and NG-RANSAC~\cite{Brachmann2019NG_RANSAC}.
	
\begin{figure}[t]
		\centering
		\includegraphics[width=0.99\linewidth]{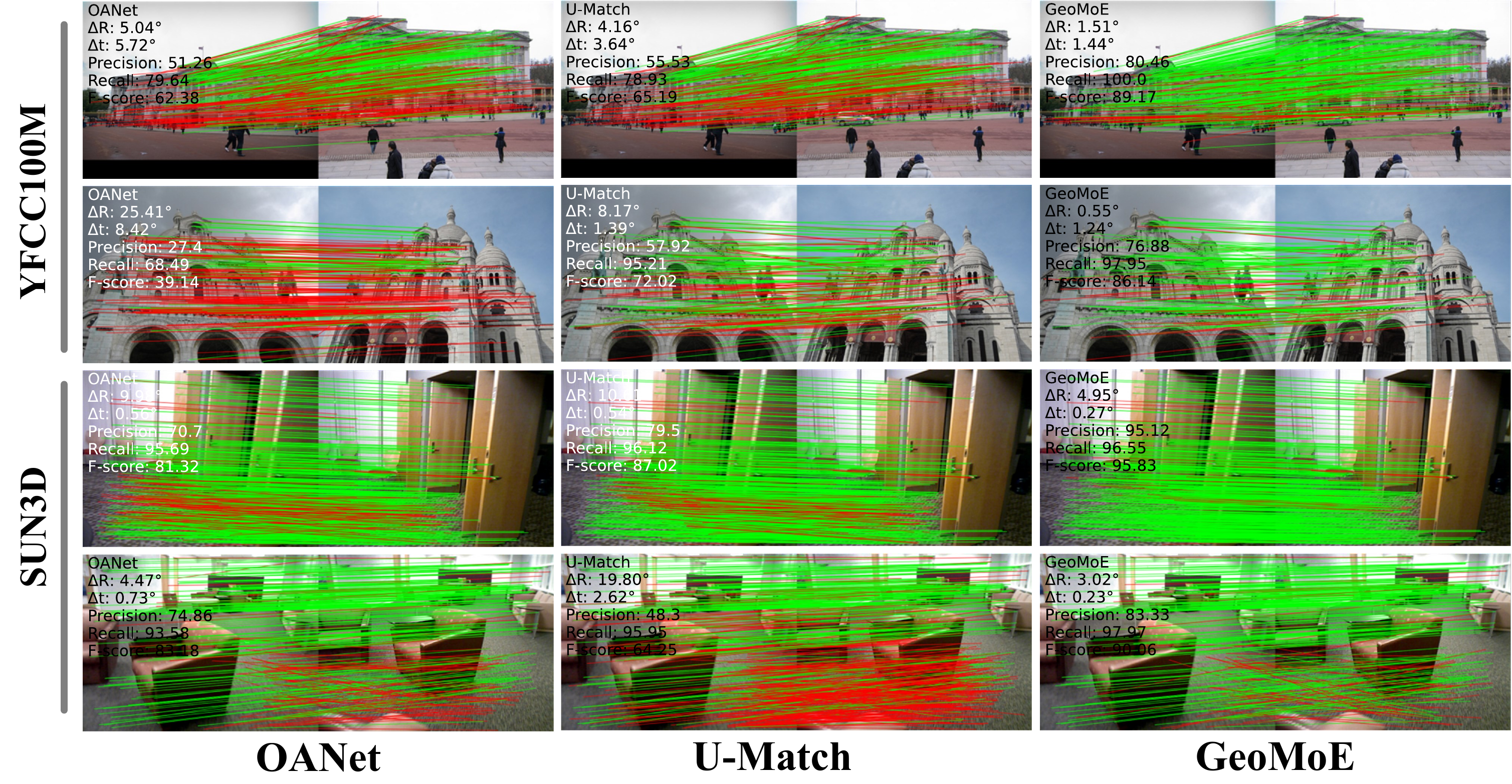}
		\vspace{-0.05in}
		\caption{Qualitative comparison of outlier rejection. From left to right: matching results of OANet, U-Match, and GeoMoE. Green and red lines denote correct and incorrect matches, respectively. Estimated relative poses are shown in the top-left corner. Zoom in for better visibility.}
		\label{pose}
	\end{figure}

	\noindent{\textit{Results:}} All quantitative evaluation results are reported in Table \ref{table1:Relative Pose Estimation}. As shown, GeoMoE consistently delivers state-of-the-art performance across most evaluation thresholds in both indoor and outdoor scenarios, underscoring its empirical effectiveness and clear advantage over existing methods.
	Notably, under the $5^\circ$ threshold, our method achieves a $10.75\%$ improvement over DeMatch when integrated with the weighted eight-point algorithm. This result highlights the effectiveness and strong potential of employing MoE to assign individual sub-fields to customized experts for fine-grained motion modeling. In addition, qualitative results in Fig. \ref{pose} show that GeoMoE can effectively suppress outliers and preserve inliers under challenging conditions such as large viewpoint changes and significant depth disparities.

	\begin{table}[t]
		\centering
		\renewcommand{\arraystretch}{1.20}
		\resizebox{\linewidth}{!}{
			\begin{tabular}{lcccc}
				\toprule
				\multirow{2}{*}{Matcher}&\multirow{2}{*}{Filter}&\multicolumn{3}{c}{YFCC100M (outdoor) (\%$\uparrow$)} \\
				\cmidrule{3-5}
				&&AUC@$5^{\circ}$ &AUC@$10^{\circ}$ &AUC@$20^{\circ}$ \\
				\midrule
				\multicolumn{1}{l}{\multirow{2}{*}{SuperPoint+NN}}&/&6.83&15.99&29.09\\
				\multicolumn{1}{l}{}&GeoMoE&\textbf{29.61}&\textbf{48.70}&\textbf{66.06}\\
				\hdashline
				\multicolumn{1}{l}{\multirow{2}{*}{SuperPoint+SuperGlue}}&/&38.24&58.47&74.64\\
				\multicolumn{1}{l}{}&GeoMoE&\textbf{39.24}&\textbf{59.78}&\textbf{75.70}\\
				\hdashline
				\multicolumn{1}{l}{\multirow{2}{*}{LoFTR}}&/&40.42&60.97&76.77\\
				\multicolumn{1}{l}{}&GeoMoE&\textbf{41.66}&\textbf{62.58}&\textbf{78.16}\\
				\bottomrule
		\end{tabular}}
		\vspace{-0.05in}
		\caption{Compatibility with learning-based matchers on YFCC100M. AUC at angular error thresholds of $5^\circ$, $10^\circ$, and $20^\circ$ for relative pose estimation is reported. \label{Compatibility}}
	\end{table}
		
	\begin{table}[t]
		\centering
		\resizebox{\linewidth}{!}{
			\begin{tabular}{lccccc}
				\toprule
				\multirow{2}{*}{Method} & \multicolumn{3}{c}{HPatches (\%$\uparrow$)} \\
				\cmidrule{2-4}
				&Acc.@3px &Acc.@5px &Acc.@10px \\
				\midrule
				PointCN  & 38.97 / 68.97  & 51.38 / 83.10  & 65.52 / \underline{92.59} \\
				OANet & 39.83 / 68.62  & 52.41 / 82.59  & 63.10 / 91.90 \\
				LMCNet & 47.93 / 72.24  & 58.62 / 85.34  & 69.83 / \underline{92.59} \\
				UMatch  & 47.07 / 70.34  & 57.41 / 85.00  & 70.00 / 91.72 \\
				DeMatch  & 46.90 / 70.69  & 60.52 / 82.76  & 71.55 / \underline{92.59}  & \\
				DeMo  & \underline{53.79} / \underline{73.10} & \underline{63.79} / \underline{85.52} & \textbf{75.69} / \underline{92.59}  \\
				GeoMoE (Ours)  & \textbf{53.83 / 74.02}  & \textbf{64.92 / 85.59}  & \underline{74.60} / \textbf{95.20} \\
				\bottomrule
			\end{tabular}
		}
		\vspace{-0.05in}
		\caption{Comparison of homography estimation results.
			The estimation accuracy (Acc.) at different pixel thresholds is reported using DLT / RANSAC as post-processing methods. }	
		\label{HPatches}
	\end{table}

	\subsubsection{Compatibility with Learning-based Matchers}
	As a general-purpose outlier rejection method, GeoMoE can accept an arbitrary set of putative correspondences and effectively filter out false matches. Leveraging this flexibility, we integrate GeoMoE with several learning-based matchers on YFCC100M to assess its compatibility and potential to further improve matching accuracy.
	Specifically, we adopt the learning-based descriptor SuperPoint~\cite{DeTone2018SuperPoint}, paired with NN Matcher and the sparse matcher SuperGlue~\cite{Sarlin2020SuperGlue}, and the detector-free matcher LoFTR~\cite{Sun2021LoFTR}. All models are evaluated using official pretrained weights. For each image pair, $2k$ keypoints are extracted using SuperPoint. Table~\ref{Compatibility} presents relative pose estimation results with and without GeoMoE.
	As shown, GeoMoE consistently improves the pose estimation accuracy of all matchers across all evaluation thresholds, demonstrating strong compatibility and generalization capability.

	\subsection{Homography Estimation}
	\subsubsection{Datasets and Evaluation Protocols}
	We conducted homography estimation experiments on the HPatches~\cite{balntas2017hpatches} benchmark, which consists of 116 image sequences, each comprising one reference image and five query images with corresponding ground-truth homography matrices. For each image, up to $4k$ keypoints are extracted, and putative correspondences are established via an NN matcher. Homographies are estimated using the Direct Linear Transform (DLT) algorithm and RANSAC as post-processing.
	We assess performance by computing the average corner error from warping with estimated and ground-truth homographies, and report the percentage of correct estimates under 3, 5, and 10 pixel thresholds.
	
	\subsubsection{Results}
	Table \ref{HPatches} presents the quantitative results of homography estimation on the HPatches dataset. GeoMoE consistently achieves top performance across most threshold settings and post-processing strategies, clearly demonstrating its effectiveness in outlier rejection.
	
	\begin{table}[t]
		\centering
		\renewcommand{\arraystretch}{1.10}
		\resizebox{\linewidth}{!}{
			\setlength{\tabcolsep}{5mm}{
				\begin{tabular}{lccc}
					\toprule
					\multirow{2}{*}{Method} & \multicolumn{3}{c}{3DMatch~\cite{Zeng2017_3DMatch}} \\
					\cmidrule{2-4}
					& RR (\%$\uparrow$) & Prec (\%$\uparrow$) & F-score (\%$\uparrow$) \\
					\midrule
					RANSAC   & 66.77 & 65.87 & 62.24 \\
					PointCN & 73.66 & 64.60 &61.49\\
					OANet     & 76.90 & 67.91 & 68.58 \\
					DeMatch   & 77.29 &68.74 & 69.82 \\
					U-Match    & 76.30 & 67.61 & 68.64 \\
					DeMo     & 77.10 & 68.94 & 69.58 \\
					PointDSC &\underline{77.50}&\underline{69.34}&\textbf{70.68}\\
					GeoMoE (Ours)  &\textbf{78.33} & \textbf{69.65} & \underline{70.65} \\
					\bottomrule
		\end{tabular}}}
		\vspace{-0.05in}
		\caption{Results of point cloud registration on 3DMatch. \label{PointCloudRegistration}}
	\end{table}

	\subsection{Point Cloud Registration}
	Point cloud registration is a fundamental task in computer vision, aiming to estimate the optimal rigid transformation between two point clouds.
	
	\subsubsection{Datasets and Evaluation Protocols}
	We perform point cloud registration experiments on the 3DMatch dataset~\cite{Zeng2017_3DMatch}. Due to the higher dimensionality of point cloud data, all methods, except PointDSC~\cite{Bai2021PointDSC}, are retrained on 3DMatch. For evaluation, we follow the registration recall (RR) metric from PointDSC, which considers an alignment successful if the rotation error is below $20^\circ$ and the translation error is less than $30 cm$. We additionally report inlier precision (Prec) and F-score to quantify the effectiveness of outlier filtering.
	
	\subsubsection{Results}
	The quantitative results of point cloud registration are presented in Table \ref{PointCloudRegistration}. GeoMoE achieves superior overall performance, particularly excelling in registration recall, the primary evaluation metric, and notably surpasses PointDSC, a method specifically tailored for point cloud registration. These results highlight the strong cross-task generalization capability of GeoMoE, enabled by its motion field decomposition and expert-driven modeling strategy. Moreover, the visualization results are provided in the \textit{S.M}.
	
	\begin{table}[t]
		\resizebox{\linewidth}{!}{
			\centering
			\begin{tabular}{lcccccccc}
				\toprule
				\multirow{2}{*}{Method} & \multicolumn{3}{c}{YFCC100M (\%$\uparrow$)}
				&& \multicolumn{4}{c}{SUN3D (\%$\uparrow$)}\\
				\cmidrule{2-4} \cmidrule{6-9}
				& RootSIFT     & LIFT     & SuperPoint
				&& SIFT     	& RootSIFT     & LIFT & SuperPoint
				\\
				\midrule
				PointCN & 24.71 & 15.31 & 14.94 && 1.56
				& 1.71   & 1.99 & 3.52    \\
				OANet & 36.46  & 28.83 & 20.54
				&& 3.37   & 3.59 & 3.20  & 3.39    \\
				UMatch & 52.49  & 38.86 & 25.90
				&& 7.63   & 7.66 & 4.84 & 3.45     \\
				BCLNet & 47.04 & 37.88 & 26.48
				&& 5.33 & 5.50 & 3.11 & 1.96 \\
				DeMatch & 53.32 & 42.27 & \underline{29.84}
				&& 7.12   & 7.31 & 6.35 & \underline{5.18}  \\		
				DeMo & \underline{55.56} &\underline{44.23} & 28.45
				&& \underline{8.22} & \underline{8.53} & \underline{6.89}& 5.01 \\	
				GeoMoE(Ours) & \textbf{56.74} & \textbf{45.00} & \textbf{30.00}
				&& \textbf{8.52}   & \textbf{8.84} & \textbf{7.39} & \textbf{6.11} \\		
				\bottomrule
			\end{tabular}
		}
		\vspace{-0.05in}
		\caption{Generalization ability. AUC$@10^\circ$ on YFCC100M is reported using the weighted eight-point algorithm.}	
		\label{Generalization Ability}
	\end{table}

	\begin{table}[t]
		\centering
		\renewcommand{\arraystretch}{1.12}
		\resizebox{\linewidth}{!}{
			\begin{tabular}{lccccc}
				\toprule
				\multirow{2}{*}{Method}&
				\multirow{2}{*}{Size(M) ($\downarrow$)}&
				\multirow{2}{*}{Flops(G) ($\downarrow$)}&
				\multicolumn{3}{c}{YFCC100M}\\
				\cmidrule{4-6}
				
				& & & Mem(MB) ($\downarrow$) &Time(ms) ($\downarrow$) & AUC@$5^{\circ}$ ($\uparrow$) \\
				\midrule
				LMCNet    & 0.926 & 2.323 & 121.38 & 301.54&22.35\\
				ConvMatch++ & 14.01& 5.379 & 245.67 & 52.14&29.43\\
				DeMatch &5.853  & 2.346  &   109.21 & 46.16&30.89\\
				U-Match   & 7.764   & 3.740    & 144.14& 65.34&30.93\\
				BCLNet   & 4.460   & 8.971    & 141.01& 59.25&27.04\\
				DeMo  & 4.520 & 3.612 & 90.22 & 54.78&32.57\\
				GeoMoE (Ours)  & 5.552 & 3.106 & 106.73 & 56.14&34.21\\
				\bottomrule 	
		\end{tabular}}
		\vspace{-0.05in}
		\caption{Comparative results of computational usage.}
		\label{ComputationalUsage}
	\end{table}

	\subsection{Analysis}
	\subsubsection{Generalization Ability}
	Generalization performance serves as a key indicator of the robustness of a model, gauging its ability to adapt across varying datasets and descriptor types. To evaluate this, we test the model trained with SIFT on YFCC100M under alternative descriptors (RootSIFT~\cite{arandjelovic2012three}, LIFT~\cite{yi2016lift}, SuperPoint) across both YFCC100M and SUN3D. Except for SuperPoint, which yields $1k$ keypoints, all descriptors extract $2k$ keypoints. As shown in Table~\ref{Generalization Ability}, GeoMoE consistently delivers state-of-the-art results across all combinations, highlighting the remarkable adaptability and the superior generalization of its expert-driven sub-field modeling, tailored to capture localized motion dynamics.

	\subsubsection{Computation Usage}
	We systematically evaluate the computational cost of GeoMoE in comparison to existing baselines, in terms of model parameter size, FLOPs, peak memory usage (Mem), and average inference time per image on the YFCC100M dataset. As shown in Table~\ref{ComputationalUsage}, despite incorporating a Mixture-of-Experts architecture, well known for increasing parameter counts, GeoMoE maintains a compact computational profile while achieving state-of-the-art performance. These results demonstrate that GeoMoE is both computationally efficient and deployment-friendly.

	\subsubsection{Visualization of MoE Behavior Across Sub-Fields}
\begin{figure}[t]
		\centering
		\includegraphics[width=0.99\linewidth]{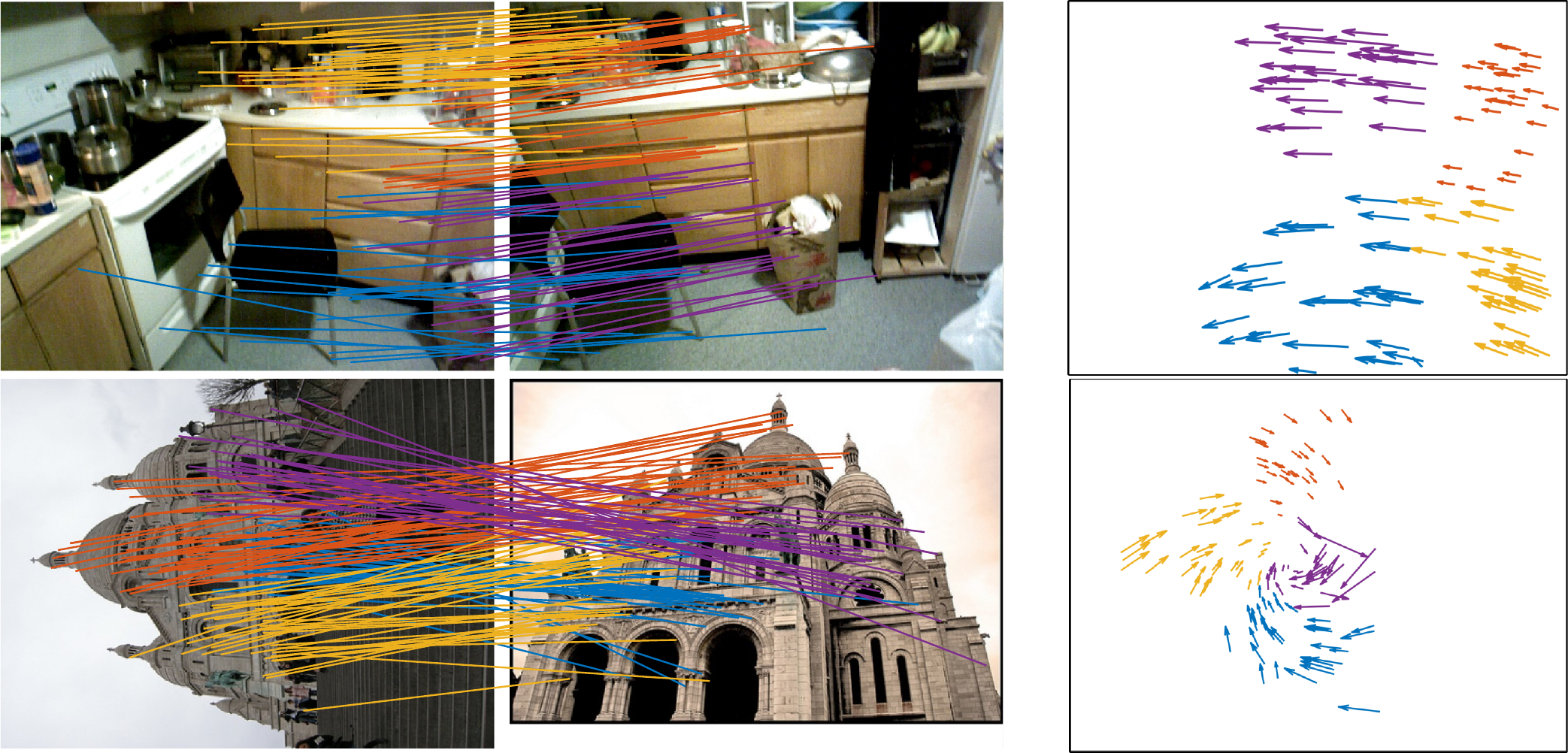}
		\vspace{-0.05in}
		\caption{Visualization of expert assignments in F-MoE.}
		\label{visual_moe}
	\end{figure}

	To offer a clearer insight into the functionality of F-MoE, Fig.~\ref{visual_moe} visualizes the matches and their corresponding motion vectors for one representative sub-field assigned to each expert. For each expert, a sub-field is randomly selected from its response region according to the expert response probabilities $\mathcal{R}$ defined in Eq.~\ref{Routing}. Distinct colors indicate assignment to different customized experts. As illustrated, each sub-field exhibits a characteristic motion pattern, attending to different objects (e.g., first row) or spatial regions (e.g., second row), and is delegated to the most suitable expert for tailored modeling and refinement.
	
	\subsubsection{Ablation Studies}
	We conduct ablation experiments on the YFCC100M dataset, with results reported in Table \ref{Ablation}. The last row represents the complete GeoMoE model. As observed from rows (i)-(v), incorporating probability injection during the decomposition process, bi-path enhancement, and F-MoE for sub-field-specific modeling in MBPR each contributes to incremental performance gains, collectively validating the effectiveness of our designs.
	
	More analytical experiments are provided in the \textit{S.M.}

	\begin{table}[t]
		\centering
		\renewcommand{\arraystretch}{1.1}
		\resizebox{\linewidth}{!}{
			\begin{tabular}{ccccccccc}
				\toprule
				Num. & Prob. & S-Path. & C-Path. & F-MoE & @5$^\circ$ & @10$^\circ$ & @20$^\circ$ \\
				\midrule
				(i)   &  &  &  & & 31.69 & 53.65 & 71.24 \\
				(ii)  & \checkmark & & & & 32.37 & 53.65 & 71.16 \\
				(iii) & \checkmark &  & & \checkmark &32.78 & 54.60 & 72.16 \\
				(iv)  &\checkmark& \checkmark &             &\checkmark & 33.14 &55.31 &72.89 \\
				(v)   & \checkmark    & \checkmark  & \checkmark &   &32.59 &54.92 & 72.58 \\
				\hdashline
				(vi)  & \checkmark & \checkmark & \checkmark & \checkmark &\textbf{34.21}& \textbf{56.25}&\textbf{73.53} \\
				\bottomrule
		\end{tabular}}
		\vspace{-0.05in}
		\caption{Ablation study. AUC on YFCC100M with the weighted eight-point algorithm are shown. ``Prob.'' denotes probability injection in PPGD. ``S-Path'' and ``C-Path'' represent the spatial and channel paths in MBPR, respectively.}
		\label{Ablation}
	\end{table}

	\section{Conclusion}
	In this work, we propose GeoMoE, a novel and unified framework for two-view geometry that harnesses a Mixture-of-Experts paradigm to model diverse and heterogeneous motion patterns, particularly under extreme viewpoints, large-scale variations, and significant depth disparities. GeoMoE incorporates a probabilistic prior-guided decomposition strategy to decompose the motion field into sub-fields, and an MoE-enhanced bi-path rectifier to assign each sub-field to customized experts for precise modeling. With its streamlined architecture, GeoMoE achieves state-of-the-art performance while maintaining high computational efficiency, enabling broad applicability in real-world scenarios.
	
	
	\bibliography{geomoe}
	\clearpage
	\twocolumn[
	\begin{center}
		{\LARGE \bf GeoMoE: Divide-and-Conquer Motion Field Modeling with Mixture-of-Experts for Two-View Geometry \\[0.5em]}
		{\large Supplementary Material}
	\end{center}
	\vspace{1em}
	]
	
	We provide the training details for GeoMoE, along with experimental settings and more visualization results. Furthermore, we conduct extended analyses to gain deeper insights into the GeoMoE architecture.
	\section{Implementation Details}
	\subsection{Training details}
	Training is performed using the Adam optimizer with an initial learning rate of $10^{-4}$ for the first $80k$ iterations. The learning rate then decays exponentially until reaching $500k$ iterations. The batch size is set to $32$. All training and evaluation experiments are conducted on NVIDIA RTX $3090$ and RTX $4090$ GPUs.
	\section{Experiments}
	\subsection{Relative Poae Estimation}
	\subsubsection{Comparisons with Outlier Rejection Methods.}
	YFCC100M~\cite{thomee2016yfcc100m} contains 100 million outdoor images divided into 72 sequences, 68 of which are used for training and validation, following the same split as OANet~\cite{zhang2019learning}, and the rest for testing. The ground-truth matches are generated based on camera poses and 3D scene reconstructions provided by COLMAP~\cite{schonberger2016structure}. SUN3D~\cite{xiao2013sun3d} is an indoor RGB-D dataset with camera pose information, containing 254 scene sequences, of which 239 are used for training and validation, and the remainder for testing.
	For each image pair, up to $2k$ keypoints are extracted using SIFT~\cite{Lowe2004SIFT}, and putative correspondences are established via an NN matcher. 
	To estimate the relative pose, we apply the weighted eight-point algorithm and RANSAC~\cite{fischler1981random} to compute the essential matrix.
	It is worth noting that traditional methods do not directly predict the essential matrix, and thus often rely solely on RANSAC as a robust estimator. 
	
	\subsection{Point Cloud Registration}
	\subsubsection{Datasets and Evaluation Protocols}
	Following PointDSC~\cite{Bai2021PointDSC}, each point cloud pair is generated with at least $30\%$ overlap. Features are extracted using the FPFH descriptor~\cite{Rusu2009FPFH}, and $1k$ keypoints are randomly sampled. Putative correspondences are established via an NN matcher, followed by outlier rejection and robust rigid transformation estimation.

	\begin{figure}[t]
		\centering
		\includegraphics[width=1\linewidth]{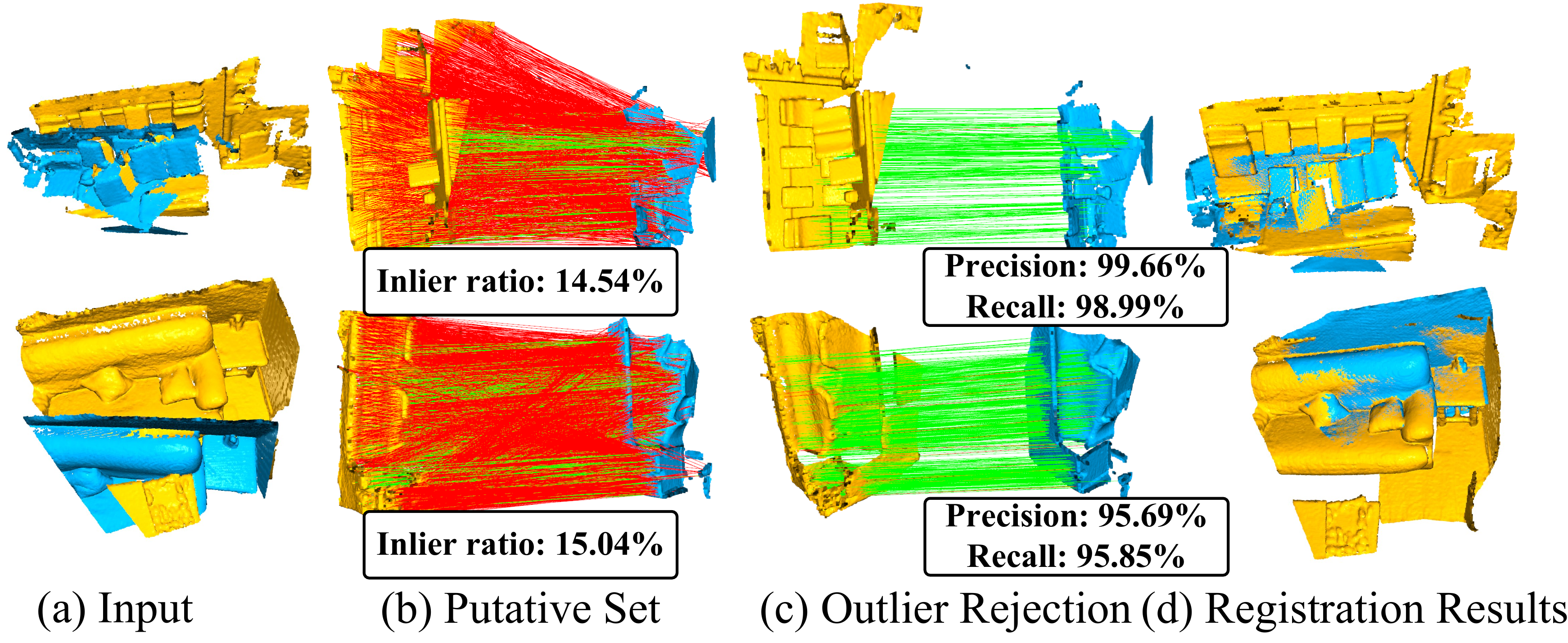}
		\caption{Visualization of point cloud registration with GeoMoE on two scenes from 3DMatch. Inlier ratio, precision, and recall are annotated in the figure.}
		\label{pcr}
	\end{figure}
	\subsubsection{Results:}
	Fig. \ref{pcr} provides visualizations of both the outlier filtering and final registration outcomes. Even under extremely low inlier ratios, our method exhibits a remarkable ability to suppress outliers while preserving reliable inliers, thereby enabling highly accurate alignment. These results demonstrate the reliability and strong generalization capability of our probabilistic prior-guided decomposition and customized expert modeling strategies.
	
	\subsection{Analysis}
	\subsubsection{Parameter Setting}
\begin{figure}[t]
		\centering
		\includegraphics[width=1\linewidth]{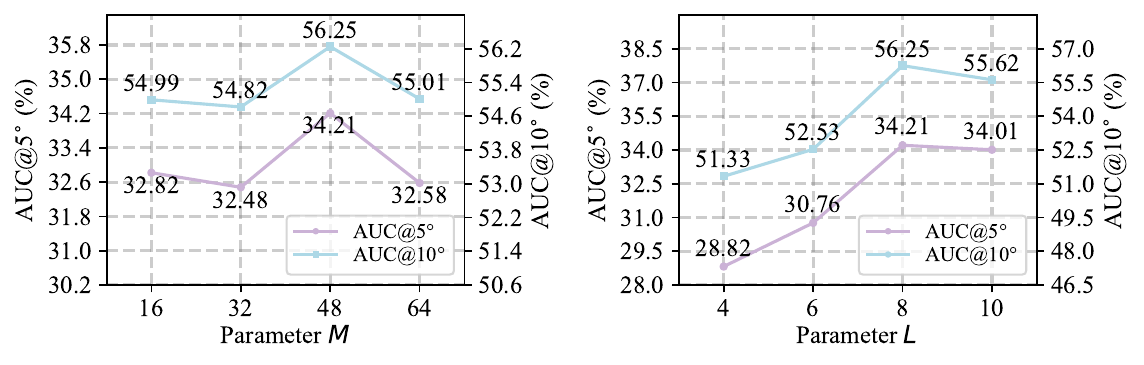}
		\caption{Parameter setting. AUC$@5^\circ$ and AUC$@10^\circ$ on YFCC100M with the weighted eight-point algorithm.}
		\label{para}
	\end{figure}

	The architectural configuration, the number of layers $L$ and sub-fields $M$, plays a pivotal role in shaping model behavior. Deeper architectures and finer sub-field granularity generally entail higher computational cost, while overly coarse sub-field divisions may undermine the model’s robustness in complex scenes. As illustrated in Fig. \ref{para}, we investigate how varying $L$ and $M$ affects performance on the relative pose estimation task. The results substantiate the effectiveness of our design choice ($L=8$ and $M=48$).

\end{document}